\documentclass{article}

\usepackage{arabtex}

\usepackage{PRIMEarxiv}

\usepackage[utf8]{inputenc} 
\usepackage[T1]{fontenc}    
\usepackage{hyperref}       
\usepackage{url}            
\usepackage{booktabs}       
\usepackage{amsfonts}       
\usepackage{nicefrac}       
\usepackage{microtype}      
\usepackage{lipsum}
\usepackage{fancyhdr}       
\usepackage{graphicx}       
\graphicspath{{media/}}     

\usepackage[figuresright]{rotating}
\usepackage{times}
\usepackage{latexsym}
\usepackage{multirow}
\usepackage{lineno}
\usepackage{tabularx}
\usepackage{array,booktabs}
\usepackage{etoolbox}
\usepackage{microtype}
\usepackage{utf8}
\usepackage{graphicx}
\usepackage{verbatim}
\usepackage{utf8}
\usepackage{pdfpages}
\usepackage{float}
\usepackage[cmex10]{amsmath}
\usepackage{amssymb}
\usepackage{epstopdf}
\usepackage{amsthm,caption}

\renewenvironment{abstract}%
         {\centerline{\large\bf Abstract}%
          \begin{list}{}%
             {\setlength{\rightmargin}{0.6cm}%
              \setlength{\leftmargin}{0.6cm}}%
           \item[]\ignorespaces}%
         {\unskip\end{list}}

\title{AraCOVID19-MFH:  Arabic COVID-19 Multi-label Fake News and Hate Speech Detection Dataset
}

\author{
  Mohamed Seghir Hadj Ameur, Hassina Aliane \\
  Research and Development in Digital Humanities Division \\
  Research Centre on Scientific and Technical Information (CERIST) \\
  Algiers, Algeria\\
  \texttt{\{mhadjameur, ahassina\}@cerist.dz} \\
}

\begin{document}
\setcode{utf8}
\maketitle

\begin{abstract}
Along with the COVID-19 pandemic, an “infodemic” of false and misleading information has emerged and has complicated the COVID-19 response efforts. Social networking sites such as Facebook and Twitter have contributed largely to the spread of rumors, conspiracy theories, hate, xenophobia, racism, and prejudice.  To combat the spread of fake news, researchers around the world have and are still making considerable efforts to build and share COVID-19 related research articles, models, and datasets. This paper releases “AraCOVID19-MFH”\footnote{\url{https://github.com/MohamedHadjAmeur/AraCOVID19-MFH}} a manually annotated multi-label Arabic COVID-19 fake news and hate speech detection dataset. Our dataset contains 10,828 Arabic tweets annotated with $10$ different labels. The labels have been designed to consider some aspects relevant to the fact-checking task, such as the tweet's check worthiness, positivity/negativity, and factuality.  To confirm our annotated dataset’s practical utility, we used it to train and evaluate several classification models and reported the obtained results. Though the dataset is mainly designed for fake news detection, it can also be used for hate speech detection, opinion/news classification, dialect identification, and many other tasks.
\end{abstract}


\section{Introduction}
Coronavirus disease (COVID-19) is an infectious respiratory disease caused by the “Sars-CoV-2” virus \cite{lai2020severe}. It was discovered in Wuhan, China, in December 2019, and declared a global pandemic by the World Health Organization (WHO) in March 2020 \cite{di2020coronavirus}. To reduce the spread of the virus, governments have adopted several measures such as closing borders, travel restrictions, quarantine, and containment. As of late January 2021, COVID-19 has caused more than 100 million confirmed cases and 2 million deaths worldwide\footnote{\url{https://www.worldometers.info/coronavirus/}}. The World Health Organization has reported that along with the COVID-19 pandemic, an “infodemic” of false and misleading information has emerged and has complicated the COVID-19 response efforts\footnote{\url{https://www.who.int/news-room/feature-stories/detail/immunizing-the-public-against-misinformation}}. Indeed, with over 4.2 million active users\footnote{\url{https://datareportal.com/social-media-users}}, social networking sites such as Facebook and Twitter have contributed largely to the spread of rumors, conspiracy theories, hate, xenophobia, racism, and prejudice. The spread of these misinformations has led people to distrust some medical treatments and vaccines and to even refuse several essential COVID-19 protection measures such as self-isolation, social distancing, and wearing masks \cite{islam2020covid, Hakak9129700, bridgman2020causes}.
To combat the spread of misinformation, many organizations around the world such as PolitiFact\footnote{\url{https://www.politifact.com/}}, FactCheck\footnote{\url{https://www.factcheck.org/}}, and Snopes\footnote{\url{https://www.snopes.com/}} have made considerable efforts to verify news articles and social media publications. However, those organizations’ manual efforts are not enough to cover and fact-check all the news and posts that are being shared on a daily basis. Thus, the need for automatic fake news detection to ease the burden on human fake-news annotators. The task’s main goal is to automatically evaluate the degree of truthfulness/trustworthiness of a given claim or news. Of course, addressing this task requires solving several challenges, such as identifying the factual articles that can be judged as fake or real, estimating their fact-checking worthiness, and assessing their content in terms of hate, racism, threats, etc.
The quality of fake news detection models is heavily dependant on the size and richness of the datasets upon which they are trained. Thus, tremendous efforts are continuously being made to create several annotated datasets for the task of fake news detection in the context of the new emerging COVID-19 pandemic \cite{Cui_2020arXiv200600885C, Gautam_2006.11343, Zhou_10.1145/3340531.3412880}. However, for many low-resource languages, such sophisticated datasets are not available or are not rich enough.
In this paper, we release “AraCOVID19-MFH”\footnote{\url{https://github.com/MohamedHadjAmeur/AraCOVID19-MFH}} a manually annotated multi-label Arabic COVID-19 fake news and hate speech detection dataset. Our dataset contains 10,828 Arabic tweets annotated with $10$ different labels. The labels have been designed to consider some aspects relevant to the fact-checking task, such as the tweet's check worthiness, positivity/negativity,  dialect, and factuality. Though the dataset is mainly designed for fake news detection, it can also be used for hate speech detection, opinion/news classification, dialect identification, and many other tasks. To confirm our annotated dataset's practical utility, we used it to train and evaluate several classification models and reported the obtained results. 
To the best of our knowledge, there are no Arabic COVID-19 fake news detection datasets that are as large and as rich as the one we are releasing in this paper.

The remainder of this paper is organized as follows: Section \ref{Related_Work} presents the fake news detection datasets that have been published in the context of the COVID-19 pandemic. The details of our dataset collection, construction, and statistics are then provided in Section \ref{Dataset_section}. Then, in Section \ref{Exp_setup_section}, we present and discuss the tests we have done and the results we have obtained. Finally, In Section \ref{Conclusion}, we conclude our work and highlight some possible future improvements.

\section{Related Work}
\label{Related_Work}


Though the coronavirus pandemic appeared only a year ago, it has received a considerable amount of attention from the research community. In this section, we will start by highlighting some datasets that have been released to combat the spread of COVID-19, then we will summarize the ones that are most relevant to the task of fake news detection\footnote{For a detailed survey of fake news detection studies, we refer the readers to \cite{zhou2020survey,Shu1011453137593137600,thorne-vlachos-2018-automated}.}.
Since the pandemic occurred at the end of December 2019, considerable efforts have been made to build large COVID-19 datasets.  For instance, 
Wang et al. \cite{wang-etal-2020-cord} published COVID-19 Open Research Dataset (CORD-19), a dataset containing more than 128,000 scientific papers regarding COVID-19. 
Kleinberg et al. \cite{kleinberg2020measuring}
released Real World Worry Dataset (RWWD), an annotated COVID-19 emotion dataset containing 5000 English tweets along with their ground truth sentiment labels. 
Banda et al. \cite{banda2020large}
released a large-scale dataset containing 383 million tweets related to COVID-19 gathered from January to June 2020. The dataset includes raw tweets in many languages with a predominance of English, French, and Spanish.
Alqurashi et al. \cite{ArabicTwitterDataset_2004.04315}
released a dataset containing $3.9$ million raw multi-dialect Arabic tweets regarding the COVID-19 pandemic. Their dataset was collected starting from January 2020 and is still being updated periodically. 
The area of COVID-19 fake news detection has also received much attention from the research community, and multiple datasets have been released\footnote{For a detailed review of the textual datasets that have been shared in regards to the COVID-19 pandemic, we refer the readers to \cite{shuja2020covid}.}. For instance, 
Limeng and Dongwon \cite{Cui_2020arXiv200600885C} created “CoAID” an English fake news detection dataset including 4,251 news articles and 926 social network posts about COVID-19 along with their ground truth labels. 
Elhadad et al. \cite{Elhadad_10.1007/978-3-030-57796-4_25} published “COVID-19-FAKE,” an automatically annotated English and Arabic tweets dataset, collected from February 4 to March 10, 2020, and annotated using 13 machine learning algorithms and seven feature extraction techniques. 
Gautam and Durgesh \cite{Gautam_2006.11343} created “FakeCovid,” a multilingual dataset gathered from 105 countries and included a total of 40 languages. The dataset contains 5,182 verified COVID-19 news articles mostly written in English. The articles were collected between April and May 2020 from 92 different fact-checking websites and have been manually annotated into 2 categories “false” and “others”\footnote{The authors used the class “others” to group all the fact-checked articles that have not been classified as “false”.}.
Zhou et al. \cite{Zhou_10.1145/3340531.3412880} published “ReCOVery”, a repository designed and built to facilitate the research studies regarding COVID-19 disinformation detection. They collected a total of 2,029 news articles published between January and May 2020, as well as 140,820 retweets of those articles that reveal how the original articles spread on Twitter. 
Firoj et al. \cite{Firoj_2005.00033} designed a multi-label twitter dataset for disinformation detection in the context of the COVID-19 pandemic for both the English and Arabic languages. There proposed dataset has been annotated using seven questions that investigate different aspects of each tweet, for instance, the questions were designed to check if the tweet contains verifiable or false information, if it has an effect on the general public, if it needs fact-checking, etc.
Haouari et al. \cite{FatimaHaouari2010.08768} presented “ArCOV19-Rumors”, an Arabic COVID-19 Twitter dataset for misinformation detection containing 9.4K tweets. The tweets were manually annotated based on verified claims gathered from popular fact-checking websites.
Parth et al. \cite{Parth_2011_03327} manually annotated a COVID-19 fake news detection dataset containing 10,700 English social media posts and articles. Their dataset contains two ground truth labels, which are “real” and “fake”.
Tammana et al. \cite{hossain2020covidlies} published “COVIDLIES,” a binary misinformation dataset containing 6761 annotated tweets built using 86 different known COVID-19 fake news articles.
Alsudias and Rayson \cite{alsudias-rayson-2020-covid} collected 1 million Covid-19 related Arabic tweets from Twitter and analyzed them for three different tasks: (1) topic identification, (2) rumors detection, and (3) tweets' sources prediction. For the rumors detection task, they manually annotated a total of 2,000 tweets using binary labels and used several machine learning algorithms to evaluate the quality of their resulting dataset.

The related study that has followed a multi-label classification similar to ours is the one published by Firoj et al. \cite{Firoj_2005.00033}. However, unlike their dataset which considers $7$ questions and includes a very limited number of instances (around 100-200 instances per class), our dataset includes $10$ tasks and significantly more annotated instances (thousands of instances per class). To the best of our knowledge, there are no Arabic COVID-19 multi-label fake news detection datasets that are as large and as rich as the one we are releasing in this paper.




\section{Dataset}
\label{Dataset_section}
This section will first present the “AraCOVID19-MFH” dataset, its design goal, and the different labels it contains. Then it explains the process of tweets collection and annotation and provides the dataset’s statistics.

\subsection{“AraCOVID19-MFH” Dataset Description}
\label{Dataset_Description}
The majority of the existing fake news detection datasets contain only two classes, “real” and “fake” \cite{zhou2020survey}. Though this kind of dataset is far easier to build, their use cases in real-world scenarios are limited. One of their main limitations is caused by the non-factual news, which can not be classified as “real” or “fake”. Another limitation of their utility is the fact that they do not take into consideration several important aspects. For instance, they do not consider the fact-checking worthiness of the tweet (or news), which can be very important in prioritizing the urgent (more harmful) news for manual fact-checking.     

Our dataset uses multi-label classes (tasks) that consider multiple aspects of each tweet, thus increasing its utility and allowing more accurate Arabic fake news detection models to be built. It considers a total of $10$ labels (tasks) described in Table \ref{tab:classification_tweets}. One of the essential labels for fake news detection is the “Factual” label. It allows the distinction between tweets that can be classified as “real” or “fake” and the tweets that contain only personal opinions or thoughts, thus, not being verifiable. Another important label is “Worth fact-checking”, which considers the level of danger (or harm) that can be inflicted by the tweet, according to which the tweet can be prioritized for manual fact-checking.  We note that our dataset's tasks were designed so that they can be used both simultaneously and independently depending on the desired task.

\begin{table*}[h]
    \scriptsize
	\centering
	\caption{“AraCOVID19-MFH” tasks, values, and their signification}
	{\def\arraystretch{1.7}\tabcolsep=1.7pt	
		\begin{tabular}{>{\centering\arraybackslash} m{25mm} >{\centering\arraybackslash}m{22mm} >{\centering\arraybackslash} m{95mm}}\hline
			\textbf{Tasks} & \textbf{Values} & \textbf{Explanation} \\\hline
			\textbf{Contains hate} & Yes, No, Can't decide & Whether the tweet contains hate, racism, or offensive speech. \\\hline
			\textbf{Talk about a cure} & Yes, No, Can't decide & Whether the tweet contains any information or discussion about a cure, a vaccine, or other possible COVID-19 treatments. \\\hline
			\textbf{Give advice} & Yes, No, Can't decide & Whether the tweet tries to advise people or government institutions. 
			\\\hline
			\textbf{Rise moral} & Yes, No, Can't decide & Whether the tweet contains encouraging, helpful, and positive speech. 
			\\\hline
			\textbf{News or opinion} & News, Opinion, Can't decide & News: if the tweet report news or a fact. \newline Opinion: if the tweet expresses a person's opinion or thoughts. 
			\\\hline
			\textbf{Dialect} & $\ \ \ \ $MSA, \newline North Africa, \newline Middle East, \newline Can't decide$\ \ $ & Weather the tweet is written in Modern Standard Arabic (MSA), North African dialect, or Middle Eastern dialect.
			\\\hline
			\textbf{Blame and negative speech} & Yes, No, Can't decide & Whether the tweet contains blame, negative, or demoralizing speech. 
			\\\hline
			\textbf{Factual} & Yes, No, Can't decide & Whether the tweet contains information that can be verified and classified as Fake or Real.
			\\\hline
			\textbf{Worth fact-checking}  & Yes, Maybe, No, Can't decide & Whether the tweet contains an important claim or dangerous content that maybe be of worth for manual fact-checking. 
			\\\hline
			\textbf{Contains fake information}  & Yes, Maybe, No, Can't decide & Whether the tweet contains any fake information.
			\\\hline
		\end{tabular}
	}
	\label{tab:classification_tweets}
\end{table*}	

As shown in Table \ref{tab:classification_tweets}, the annotation task involves a set of labels (tasks), each one of them investigates a given aspect of the Arabic tweet.  In the following, we will try to provide more details about the dataset.

\begin{itemize}
	\item For the dialect label, we considered only three major values: Modern Standard Arabic (MSA), North African, and Middle Eastern dialects. North African dialect (also known as Maghrebi Arabic) includes all the Arabic dialects spoken in Morocco, Algeria, Tunisia, Libya, Western Sahara, and Mauritania\footnote{\url{https://en.wikipedia.org/wiki/Maghrebi_Arabic}}. Middle Eastern dialect (also known as Mashriqi Arabic) includes all the Arabic dialects spoken in the Mashriq countries\footnote{\url{https://en.wikipedia.org/wiki/Mashriqi_Arabic}}. This dialect classification is adopted to make the manual annotation much easier for the annotators who may not be very familiar with the little differences between Arabic countries’ dialects.
	
	
	
	\item Both the fake information and the worth fact-checking classes are considered only when the tweets are factual. Thus, if the tweet is not factual, we can not talk about its fact-checking worthiness nor about it being “fake” or “real”.
	
	\item We note that for all the tasks of our dataset, an additional label “Can’t decide” is added to give the annotators a choice to not make a decision for a given class. For example, if the annotator cannot decide whether the tweet is fake or not, he/she can choose the “can’t decide” label for it.
\end{itemize}

\begin{figure}[h]
	\centering
	\captionof{table}[foo]{Example of some Arabic tweets along with their respective multi-label annotations}
	\includegraphics[width=0.95\linewidth]{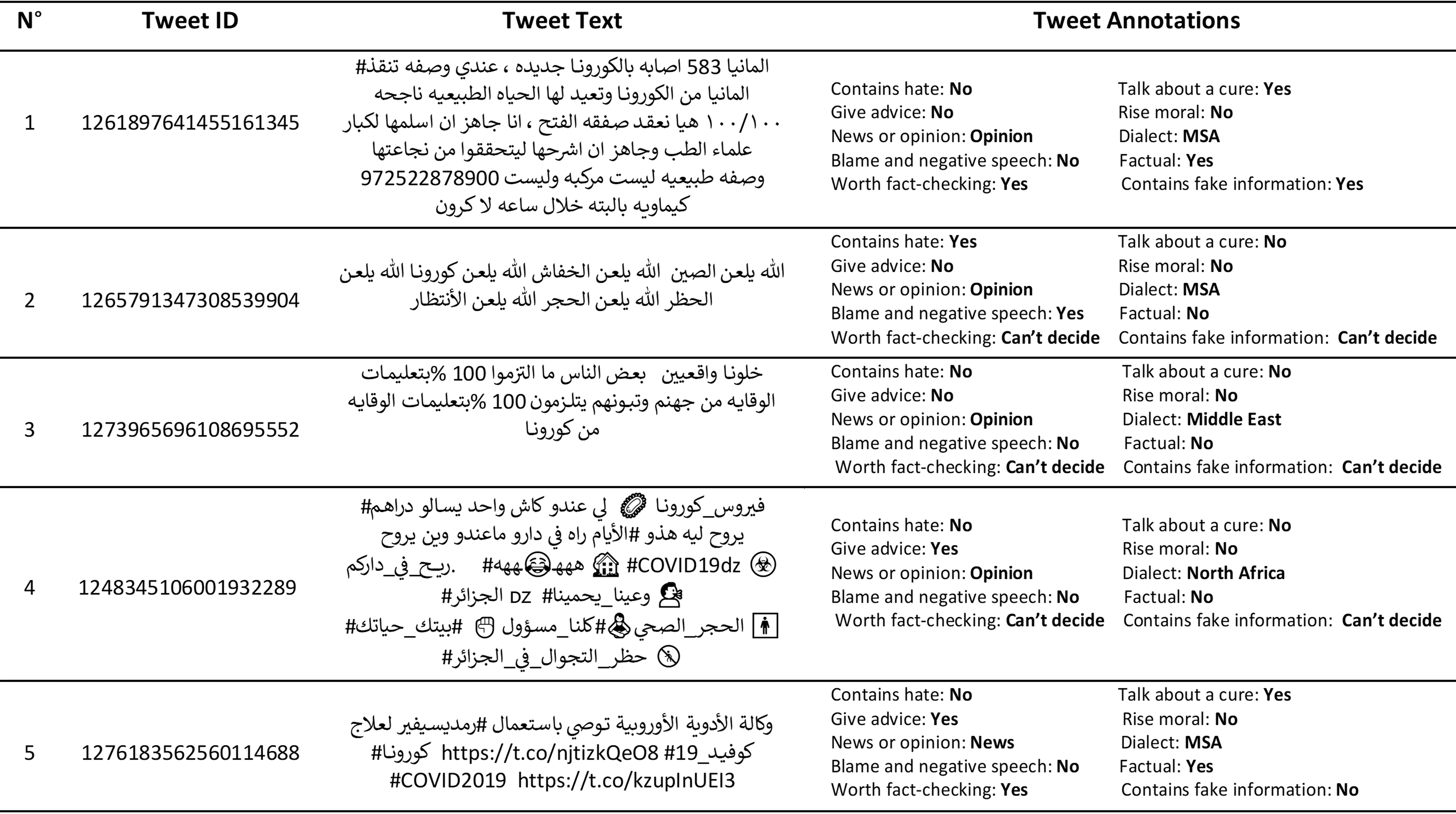}
	\label{img:exemple_tweets}
\end{figure}

Table \ref{img:exemple_tweets} shows some examples of annotated Arabic tweets.
The tweets N° 2, 3 and 4 (Table \ref{img:exemple_tweets}) are not factual. Thus they can not be classified as “fake” or “real”, unlike the tweets N° 1 and 5, which are factual. 
We note that our dataset’s released version follows the same structure as shown in Table \ref{img:exemple_tweets}. However, it does not provide the tweets’ texts (due to the Twitter regulations\footnote{\url{https://help.twitter.com/en/rules-and-policies/twitter-rules}}), instead, it only provides the tweets’ IDs that can be used to retrieve the full information of each tweet.

\subsection{Data Collection}
The first step that we followed to build the dataset was to prepare a set of keywords for each specific task, then we retrieved the tweets based on those keywords.

Portions of the keywords' lists that we prepared for each task are shown in Table \ref{table:Preclassified_topics}.

\begin{table*}[h]
	\scriptsize
	\centering
	\caption{Some task-specific keywords that have been used to collect the Arabic tweets}
	{\def\arraystretch{1.5}\tabcolsep=1.5pt	
			\begin{tabular}{cc}
				\hline
				\textbf{Topics}                                & \textbf{Keywords}                                                                                                                                                            \\ \hline
				\textbf{Base   Corona} &  \<  فيروس، الفيروس،  الوباء، وباء،   كورونا، كوفيد، الكورونا، الكوفيد                                                                                                            >\\
				\textbf{Contains   hate}                       & \< فاسدة، عصابة، كلاب، عملاء، يلعن،   حيوان، خائن خونة                                                                                                                          >\\
				\textbf{Talk   about a cure}                   & \< موديرنا، فايزر، سبوتنيك، رمديسيفير، وصفة، علاج لقاح، كلوروكين، أزيثروميسين، لقاحات، دواء                                                                                     >\\
				\textbf{Give   advice}                         & \< لكي تتجنب، لكي لا تصاب، لكي   تتفادى، يجب غسل، لحماية نفسك، لتجنب العدوى، نتفاداو، نحمو                             > \\
				\textbf{Rise   moral}                          & \< لاداعي للخوف، الصبر، اصبرو، لا تخافو، لا تقلقو، لا تستعجلو، ستزول، سيزول، سيختفي، سترجع الحياة، ماتتقلقوش، ماتخافوش                                           >\\
				\textbf{News}                                  & \< إعلان، أعلن، أعلنت، سجل، تسجيل، سجلت تناقلت، نقلت، إصابة، عاجل                                                                                                               >\\
				\textbf{Opinion}                               & \< أنا، رأيي أرى، أنتم، نحن، هم،   أظن، أشك، أعلم ظننت، أخبرتكم، قلتلكم، قلت، قلنا، قالو                                                                                        >\\
				\textbf{Blame   and negative speech \space\space}           & \< سنموت، انتهت حياتنا، حياة تعيسة، حياة كئيبة، حياة مملة، حياة سيئة، لا يوجد علاج،       ستهلكون، ستموتون، لن تعيشو، مكاش دوا >	\\ \hline
			\end{tabular}
			}
			\label{table:Preclassified_topics}
		\end{table*}

The “Base Corona” category (Table \ref{table:Preclassified_topics}) contains the COVID-19 specific keywords that we add to each task’s keywords to ensure that the retrieved tweets are related to the COVID-19 pandemic. 

The retrieved tweets were filtered in the following way:
\begin{itemize}
	\item All the retweets of a given tweet were removed.
	\item Identical tweets that share the exact same textual content (when ignoring the tweets' links) were removed. This is done to ensure that the text of each considered tweet is unique.
	\item Very short tweets that contain less than $5$ Arabic words were filtered.
	\item Tweets were gathered within the time period spanning from December 15, 2019, and December 15, 2020.
\end{itemize}
After the filtering step, we have ended up with a total of 300k unique Arabic tweets related to the COVID-19 pandemic.

\subsection{Data Annotation}
Due to the high cost of the annotation task, we only required each tweet to be annotated by one expert annotator. This allowed us to annotate a total of $10,828$ Arabic tweets from the 300k gathered tweets. We plan to perform a second annotation phase (re-annotation) at a later time in which each existing tweet will be further examined and annotated by at least two other annotators, and the remaining non annotated tweets will be gradually annotated. 
The manual annotation task was carried-out while taking into account the following guidelines and instructions:
\begin{itemize}
	\item For each tweet, the annotator has been provided with the full text of the tweet, including the links, and asked to read the tweet, check the tweet's links if necessary, and annotate it for each one of the $10$ labels (tasks). This results in a dataset in which each tweet is labeled for each one of the $10$ tasks (as shown in Table \ref{img:exemple_tweets}).
	\item For the “Contains fake information” label, the annotator needs to verify the validity of the tweet's claims or news from trusted online sources. 
	\item The dialect classification is based solely on the tweet’s text. We do not use the tweets’ geolocalisation data because they are not always correct, and also because we believe that the distinction between the considered “MSA”, “North Africa”, and “Middle East” can be easily made.
	\item If a tweet contains both news and opinion parts, the annotator needs to make the classification based on the more significant part, or he can choose the “Can't decide” value. 
	\item For the “Contains hate” label, we consider hate in a global manner, which includes racism and any type of offensive speech.
	\item We give the annotator a choice to not make a decision for any given task by selecting the value “Can’t decide”. The tweet will then be unlabelled for that specific task.  
\end{itemize}


\subsection{“AraCOVID19-MFH” Statistics}
The statistics of our “AraCOVID19-MFH” dataset are provided in Table \ref{table:stats}.


\begin{table*}[h]
    \scriptsize
	\centering
	\caption{Statistics about the number of tweets in each topic}
	{\def\arraystretch{1.7}\tabcolsep=1.7pt	
		\begin{tabular}{cc}
            \hline
            \textbf{Topics}                                 & \textbf{Tweets}                                                                       \\ \hline
            \textbf{Contains   hate}                        & \textbf{Yes}: 1233  \space\space\space    \textbf{No}: 9571  \space\space\space   \textbf{Can't   decide}: 24                               \\
            \textbf{Talk   about a cure}                    & \textbf{Yes}: 1510  \space\space\space   \textbf{No}: 9318  \space\space\space   \textbf{Can't   decide}: 0                                 \\
            \textbf{Give   advice}                          & \textbf{Yes}: 1593   \space\space\space  \textbf{No}: 9235   \space\space\space  \textbf{Can't   decide}: 0                                 \\
            \textbf{Rise   moral}                           & \textbf{Yes}: 1223   \space\space\space  \textbf{No}: 9605   \space\space\space  \textbf{Can't   decide}: 0                                 \\
            \textbf{News   or opinion}                      & \textbf{News}: 3308   \space\space\space  \textbf{Opinion}: 7520  \space\space\space   \textbf{Can't decide}: 0                             \\
            \textbf{Dialect}                                & \textbf{MSA}: 5132  \space\space\space      \textbf{North Africa}: 1346   \space\space\space \textbf{Middle East}: 4109  \space\space\space  \textbf{Can't   decide}: 241 \\
            \textbf{Blame   and negative speech}            & \textbf{Yes}: 763 \space\space\space  \textbf{No}: 10030  \space\space\space  \textbf{Can't   decide}: 35                                   \\
            \textbf{Factual}                                & \textbf{Yes}: 4480  \space\space\space \textbf{No}: 3006  \space\space\space  \textbf{Can't   decide}: 3342                                 \\
            \textbf{Worth   fact-checking}                  & \textbf{Yes}: 1335  \space\space\space \textbf{Maybe}: 2094 \space\space\space  \textbf{No}: 1158  \space\space\space  \textbf{Can't   decide}: 6241                     \\
            \textbf{Contains   fake information}            & \textbf{Yes}: 459  \space\space\space \textbf{Maybe}: 1839 \space\space\space  No: 2249  \space\space\space   \textbf{Can't   decide}: 6281                   \\
              \textbf{Total Tweets} & \textbf{10828}   \\ \hline                                                                   
        \end{tabular}
        }
        \label{table:stats}
        \end{table*}

As shown in Table \ref{table:stats}, most of the considered tasks contain more than $1000$ instances for each one of their values, which helps train robust classification models. For the “Worth fact-checking” and the “Contains fake information” tasks the number of tweets that have been annotated with the “Can't decide” value is very high because a large portion of our dataset's tweets is not factual. And as we stated at the end of section \ref{Dataset_Description}, the non-factual tweets can't be annotated for either one of those two tasks.

\section{Tests and Results}
\label{Exp_setup_section}
Our tests aim at assessing the quality of our constructed dataset and provide baseline results for each one of its tasks. To this end, for each one of our dataset's tasks, several deep learning models are trained and tested. In the following, first,  we will present the Arabic preprocessing that we performed, and the different pre-trained deep learning models that we considered. Then, we will report and discuss the results of our performed tests.

\subsection{Preprocessing}
We applied a basic preprocessing to all the collected Arabic tweets, which includes:
\begin{itemize}
	\item The removal of diacritical marks.
	\item The removal of elongated and repeated characters.
	\item Arabic characters normalization.
	\item  The removal of links and users' references (users' notifications). 
	\item Tweets tokenization in which punctuation, words, and numbers are separated. 
\end{itemize}
We note that this preprocessing have been used only when training the transformer models (Section \ref{transformer_Models}); it has not been used for the annotation task nor in the final dataset.

\subsection{Considered Models}
\label{transformer_Models}
Pretrained transformer models have been recently used in many NLP tasks and have continuously achieved new state-of-the-art results \cite{DBLP:BERT_abs-1810-04805}. In the following, we will highlight the five transformer models that we considered in our tests.

\subsubsection{Baseline Transformer Models}
In our experiments we use three pretrained transformer models:
\begin{itemize}
	\item AraBERT\footnote{\url{https://huggingface.co/aubmindlab/bert-base-arabertv02}}: A  BERT (Bidirectional Encoder Representations from Transformers) model \cite{DBLP:BERT_abs-1810-04805} pretrained on $200$ million Arabic MSA sentences gathered from different sources \cite{antoun2020arabert}. 
	\item Multilingual BERT (mBERT)\footnote{\url{https://huggingface.co/bert-base-multilingual-cased}}: A BERT-based model \cite{DBLP:BERT_abs-1810-04805} pretrained on the first $104$ major Wikipedia languages\footnote{\url{https://meta.wikimedia.org/wiki/List_of_Wikipedias}}. 
	\item Distilbert Multilingual\footnote{\url{https://huggingface.co/distilbert-base-multilingual-cased}}: A distilled version of the mBERT (smaller version), trained using knowledge distillation \cite{sanh2019distilbert}.
\end{itemize}

\subsubsection{COVID-19 Transformer Models}
The three aforementioned transformer models have been trained solely on MSA Arabic. Thus they may have difficulty dealing with Arabic dialects.  For this reason, we performed a further pretraining (or fine-tuning) for AraBERT and mBERT models using $1.5$ million tweets from the “Large Arabic Twitter Dataset on COVID-19” \cite{ArabicTwitterDataset_2004.04315} which contains raw multi-dialect Arabic tweets regarding the COVID-19 pandemic. This is done to make the transformer language models more familiar with the Arabic COVID-19 multi-dialect vocabulary. We named the resulting two models: “AraBERT COV19”\footnote{\url{https://huggingface.co/moha/arabert_c19}}, and “mBERT COV19”\footnote{\url{https://huggingface.co/moha/mbert_ar_c19}}. With these two modes, we end up with a total of five models for our experiments. 

\subsubsection{Transformers Weights' Fine-tuning}
Pretrained transformer models can often achieve better results when their weights are fine-tuned on the classification task \cite{sun2019fine}. For our tests, we investigate two training scenarios:
\begin{enumerate}
	\item Without fine-tuning: we train our five considered classification models without allowing their weights to be changed during the training process.
	\item With fine-tuning: we train our five considered classification models while allowing their weights to be updated (fine-tuned) to maximize the classification performance.
\end{enumerate}

\subsection{Software and Tools}
The implementation of the different models have been done using the following libraries:
\begin{itemize}
	\item Scikit-learn  \cite{scikit-learn}\footnote{\url{https://scikit-learn.org/stable/}} is a python-based  machine learning library. We used it to evaluate the performance of our models.
	
	\item Flair \cite{Flair_akbik2018coling}\footnote{\url{https://github.com/flairNLP/flair}}: is a framework for building state-of-the-art NLP models. We used it to train our classification models.
	
	\item Huggingface-transformers \cite{wolf-etal-2020-transformers}\footnote{\url{https://github.com/huggingface/transformers}}: is a framework for building and pretraining different state-of-the-art NLP models. We used it to fine-tune the pretrained transformer language models on the new Arabic COVID-19 data.
	
	\item PyTorch \cite{PyTorch_NEURIPS2019_9015}\footnote{\url{https://github.com/pytorch/pytorch}} is an open-source library designed for implementing deep neural networks. We used it as a backend for both the Huggingface-transformers and the Flair frameworks.
	
\end{itemize}

\subsection{Evaluation Methodology}
To evaluate the performance of our five considered classification models, we have used a stratified 5-fold cross-validation method. This is done by randomly partitioning the instances of each one of our dataset's tasks into 5 disjoint sets of equal size. In this five-fold cross-validation, five experiments are performed, in each one, one of the five sets is selected for testing, and the remaining four are used for training. For each experiment, the weighted F-score is calculated, and finally, the average F-score for all the five experiments (the 5-folds) is reported.

\subsection{Results and Discussion}
\label{Test_section}
Our tests examine two important aspects. First, we compare the transformer models' performance when their weights are fine-tuned on the classification task and when their weights are kept unchanged (with and without fine-tuning). The second aspect investigates the effect of the additional COVID-19 pretraining that we incorporated for the AraBERT and mBERT models. The results of the two aforementioned experiments are given in Table \ref{table:results}.


\begin{table*}[h]
	\scriptsize
	\centering
	\caption{Weighted F-score results for the considered transformer models with and without fine-tuning}
	{\def\arraystretch{1.5}\tabcolsep=1.5pt	
		\begin{tabular}{cccc|cc|ccc|cc}
			\cline{2-11}
			& \multicolumn{5}{c}{Without Fine-tuning}      & \multicolumn{5}{c}{With Fine-tuning}  \\ \cline{2-11}
			& \multicolumn{3}{c}{Baseline models} & \multicolumn{2}{c}{Pretrained Covid-19 models} & \multicolumn{3}{c}{Baseline   models} & \multicolumn{2}{c}{Pretrained Covid-19 models} \\ \cline{2-11}
			& arabert & mbert  & distilbert-multi & \textbf{arabert Cov19}        & \textbf{mbert Cov19}       & arabert  & mbert   & distilbert-multi & \textbf{arabert Cov19}        & \textbf{mbert Cov19}  \\ \cline{1-11}
			\textbf{Contains hate}       	& 0.8346  & 0.6675 & 0.7145   & \textbf{0.8649}  & 0.8492       & 0.9809   & 0.97    & 0.9736   & \textbf{0.9858}  & 0.9809   \\
			\textbf{Talk about a cure}   	& 0.8193  & 0.7406 & 0.7127   & 0.9055  & \textbf{0.9176}       & 0.99     & 0.9854  & 0.9774   & \textbf{0.9930}   & 0.9904   \\
			\textbf{Give advice} 			& 0.8287  & 0.6865 & 0.6974   & \textbf{0.9035}  & 0.8948       & 0.9793   & 0.9664  & 0.9764   & 0.9824  & \textbf{0.9862}   \\
			\textbf{Rise moral   }       	& 0.8398  & 0.7075 & 0.7049   & \textbf{0.8903}  & 0.8838       & 0.9618   & 0.9663  & 0.9618   & 0.97    & \textbf{0.9712}   \\
			\textbf{News or opinion }    	& 0.8987  & 0.8332 & 0.8099   & \textbf{0.9163}  & 0.9116       & 0.9552   & 0.9409  & 0.9529   & \textbf{0.9627}  & 0.9594   \\
			\textbf{Dialect}     			& 0.7533  & 0.558  & 0.5433   & \textbf{0.8230}   & 0.7682       & 0.9266   & 0.9137  & 0.9102   & 0.9281  & \textbf{0.9317}    \\
		\textbf{Blame and negative speech}  & 0.7426  & 0.597  & 0.6221   & \textbf{0.7997}  & 0.7794       & 0.9607   & 0.9476  & 0.9587   & \textbf{0.9653}  & 0.9633    \\
			\textbf{Factual}     			& 0.9217  & 0.8427 & 0.8383   & 0.9575  & \textbf{0.9608}       & 0.9958   & 0.9917  & 0.9925   & 0.995   & \textbf{0.9967}   \\
			\textbf{Worth fact-checking} 	& 0.7731  & 0.5298 & 0.5413   & 0.8265  & \textbf{0.8383}       & 0.9885   & 0.9824  & 0.9763   & \textbf{0.9907}  & 0.9891    \\
		\textbf{Contains fake information}  & 0.6415  & 0.5428 & 0.4743   & \textbf{0.7739}  & 0.7228       & 0.9417   & 0.9353  & 0.9288   & \textbf{0.9578}  & 0.9491    \\ \cline{1-11}                              
		\end{tabular}
		}
		\label{table:results}
	\end{table*}

Regarding the first experiment, we can observe that the models' weights fine-tuning was extremely helpful for all the tested transformer models with an improvement ranging between $0.1$ and $0.5$ f-score. This implies that fine-tuning the transformers' weights during the training phase of the classification is extremely helpful.
For the second experiment, we can see that the additional training that has been performed for the “AraBERT” and “mBERT” transformer models using $1.5$ million multi-dialect Arabic COVID-19 tweets was helpful both when the transformers' weights were and weren't updated. Indeed, we can observe that “AraBERT COVID-19” and “mBERT COVID-19” models achieved the best performance reaching more than $0.92$ f-score across all the tested tasks. This confirms that performing additional training using task-specific data for the pretrained transformer models is helpful for the classification tasks. 

The quality of the obtained results reflects the importance of having a large annotated dataset and confirms our adopted annotation schema's practical utility.

\section{Conclusion}
\label{Conclusion}
We have presented and released “AraCOVID19-MFH” the Arabic COVID-19 multi-label fake news and hate speech detection dataset. The dataset contains 10,828 Arabic tweets; each tweet is annotated with $10$ labels. The labels were designed to consider different aspects of each tweet, such as its check worthiness, positivity/negativity, dialect, factuality, etc. Though the dataset is mainly designed for fake news detection, it can also be used for hate speech detection, opinion/news classification, dialect identification, and many other tasks. All the dataset's tweets have been manually annotated and validated by human annotators. The quality of the final annotated dataset has been evaluated using several pretrained transformer models. Two transformer models have been fine-tuned using COVID-19 data and managed to achieve the best classification results across all the tested categories. Both our fine-tuned transformer models, along with the full annotated dataset, are freely available for research purposes.  
As future work, we plan to continue enriching our annotated dataset to make it larger and keep it up-to-date with the latest events and discussions that are shared on Twitter regarding the COVID-19 pandemic. 




\begin{thebibliography}{10}

\bibitem{lai2020severe}
Chih-Cheng Lai, Tzu-Ping Shih, Wen-Chien Ko, Hung-Jen Tang, and Po-Ren Hsueh.
\newblock Severe acute respiratory syndrome coronavirus 2 (sars-cov-2) and
  coronavirus disease-2019 (covid-19): The epidemic and the challenges.
\newblock {\em International journal of antimicrobial agents}, 55(3):105924,
  2020.

\bibitem{di2020coronavirus}
Francesco Di~Gennaro, Damiano Pizzol, Claudia Marotta, Mario Antunes, Vincenzo
  Racalbuto, Nicola Veronese, and Lee Smith.
\newblock Coronavirus diseases (covid-19) current status and future
  perspectives: a narrative review.
\newblock {\em International journal of environmental research and public
  health}, 17(8):2690, 2020.

\bibitem{islam2020covid}
Md~Saiful Islam, Tonmoy Sarkar, Sazzad~Hossain Khan, Abu-Hena~Mostofa Kamal,
  SM~Murshid Hasan, Alamgir Kabir, Dalia Yeasmin, Mohammad~Ariful Islam, Kamal
  Ibne~Amin Chowdhury, Kazi~Selim Anwar, et~al.
\newblock Covid-19--related infodemic and its impact on public health: A global
  social media analysis.
\newblock {\em The American Journal of Tropical Medicine and Hygiene},
  103(4):1621, 2020.

\bibitem{Hakak9129700}
S.~{Hakak}, W.~Z. {Khan}, M.~{Imran}, K.~R. {Choo}, and M.~{Shoaib}.
\newblock Have you been a victim of covid-19-related cyber incidents? survey,
  taxonomy, and mitigation strategies.
\newblock {\em IEEE Access}, 8:124134--124144, 2020.

\bibitem{bridgman2020causes}
Aengus Bridgman, Eric Merkley, Peter~John Loewen, Taylor Owen, Derek Ruths,
  Lisa Teichmann, and Oleg Zhilin.
\newblock The causes and consequences of covid-19 misperceptions: Understanding
  the role of news and social media.
\newblock {\em Harvard Kennedy School Misinformation Review}, 1(3), 2020.

\bibitem{Cui_2020arXiv200600885C}
Limeng {Cui} and Dongwon {Lee}.
\newblock {CoAID: COVID-19 Healthcare Misinformation Dataset}.
\newblock {\em arXiv e-prints}, page arXiv:2006.00885, May 2020.

\bibitem{Gautam_2006.11343}
Gautam~Kishore Shahi and Durgesh Nandini.
\newblock Fakecovid - a multilingual cross-domain fact check news dataset for
  covid-19.
\newblock 2020.

\bibitem{Zhou_10.1145/3340531.3412880}
Xinyi Zhou, Apurva Mulay, Emilio Ferrara, and Reza Zafarani.
\newblock {\em ReCOVery: A Multimodal Repository for COVID-19 News Credibility
  Research}, page 3205–3212.
\newblock Association for Computing Machinery, New York, NY, USA, 2020.

\bibitem{zhou2020survey}
Xinyi Zhou and Reza Zafarani.
\newblock A survey of fake news: Fundamental theories, detection methods, and
  opportunities.
\newblock {\em ACM Comput. Surv.}, 53(5), September 2020.

\bibitem{Shu1011453137593137600}
Kai Shu, Amy Sliva, Suhang Wang, Jiliang Tang, and Huan Liu.
\newblock Fake news detection on social media: A data mining perspective.
\newblock {\em SIGKDD Explor. Newsl.}, 19(1):22–36, September 2017.

\bibitem{thorne-vlachos-2018-automated}
James Thorne and Andreas Vlachos.
\newblock Automated fact checking: Task formulations, methods and future
  directions.
\newblock In {\em Proceedings of the 27th International Conference on
  Computational Linguistics}, pages 3346--3359, Santa Fe, New Mexico, USA,
  August 2018. Association for Computational Linguistics.

\bibitem{wang-etal-2020-cord}
Lucy~Lu Wang, Kyle Lo, Yoganand Chandrasekhar, Russell Reas, Jiangjiang Yang,
  Doug Burdick, Darrin Eide, Kathryn Funk, Yannis Katsis, Rodney~Michael
  Kinney, Yunyao Li, Ziyang Liu, William Merrill, Paul Mooney, Dewey~A.
  Murdick, Devvret Rishi, Jerry Sheehan, Zhihong Shen, Brandon Stilson, Alex~D.
  Wade, Kuansan Wang, Nancy Xin~Ru Wang, Christopher Wilhelm, Boya Xie,
  Douglas~M. Raymond, Daniel~S. Weld, Oren Etzioni, and Sebastian Kohlmeier.
\newblock {CORD-19}: The {COVID-19} open research dataset.
\newblock In {\em Proceedings of the 1st Workshop on {NLP} for {COVID-19} at
  {ACL} 2020}, Online, July 2020. Association for Computational Linguistics.

\bibitem{kleinberg2020measuring}
Bennett Kleinberg, Isabelle van~der Vegt, and Maximilian Mozes.
\newblock Measuring emotions in the covid-19 real world worry dataset.
\newblock In {\em Proceedings of the 1st Workshop on NLP for COVID-19 at ACL
  2020}, 2020.

\bibitem{banda2020large}
Juan~M Banda, Ramya Tekumalla, Guanyu Wang, Jingyuan Yu, Tuo Liu, Yuning Ding,
  and Gerardo Chowell.
\newblock A large-scale covid-19 twitter chatter dataset for open scientific
  research--an international collaboration.
\newblock {\em arXiv preprint arXiv:2004.03688}, 2020.

\bibitem{ArabicTwitterDataset_2004.04315}
Sarah Alqurashi, Ahmad Alhindi, and Eisa Alanazi.
\newblock Large arabic twitter dataset on covid-19, 2020.

\bibitem{shuja2020covid}
Junaid Shuja, Eisa Alanazi, Waleed Alasmary, and Abdulaziz Alashaikh.
\newblock Covid-19 open source data sets: A comprehensive survey.
\newblock {\em Applied Intelligence}, pages 1--30, 2020.

\bibitem{Elhadad_10.1007/978-3-030-57796-4_25}
Mohamed~K. Elhadad, Kin~Fun Li, and Fayez Gebali.
\newblock Covid-19-fakes: A twitter (arabic/english) dataset for detecting
  misleading information on covid-19.
\newblock In Leonard Barolli, Kin~Fun Li, and Hiroyoshi Miwa, editors, {\em
  Advances in Intelligent Networking and Collaborative Systems}, pages
  256--268, Cham, 2021. Springer International Publishing.

\bibitem{Firoj_2005.00033}
Firoj Alam, Shaden Shaar, Fahim Dalvi, Hassan Sajjad, Alex Nikolov, Hamdy
  Mubarak, Giovanni Da~San Martino, Ahmed Abdelali, Nadir Durrani, Kareem
  Darwish, and Preslav Nakov.
\newblock Fighting the covid-19 infodemic: Modeling the perspective of
  journalists, fact-checkers, social media platforms, policy makers, and the
  society, 2020.

\bibitem{FatimaHaouari2010.08768}
Fatima Haouari, Maram Hasanain, Reem Suwaileh, and Tamer Elsayed.
\newblock Arcov19-rumors: Arabic covid-19 twitter dataset for misinformation
  detection, 2020.

\bibitem{Parth_2011_03327}
Parth Patwa, Shivam Sharma, Srinivas PYKL, Vineeth Guptha, Gitanjali Kumari,
  Md~Shad Akhtar, Asif Ekbal, Amitava Das, and Tanmoy Chakraborty.
\newblock Fighting an infodemic: Covid-19 fake news dataset, 2020.

\bibitem{hossain2020covidlies}
Tamanna Hossain, Robert~L Logan~IV, Arjuna Ugarte, Yoshitomo Matsubara, Sean
  Young, and Sameer Singh.
\newblock Covidlies: Detecting covid-19 misinformation on social media.
\newblock In {\em Proceedings of the 1st Workshop on NLP for COVID-19 (Part 2)
  at EMNLP 2020}, 2020.

\bibitem{alsudias-rayson-2020-covid}
Lama Alsudias and Paul Rayson.
\newblock {COVID-19} and {Arabic} {Twitter}: How can {Arab} world governments
  and public health organizations learn from social media?
\newblock In {\em Proceedings of the 1st Workshop on {NLP} for {COVID-19} at
  {ACL} 2020}, Online, July 2020. Association for Computational Linguistics.

\bibitem{DBLP:BERT_abs-1810-04805}
Jacob Devlin, Ming-Wei Chang, Kenton Lee, and Kristina Toutanova.
\newblock Bert: Pre-training of deep bidirectional transformers for language
  understanding.
\newblock In {\em Proceedings of the 2019 Conference of the North American
  Chapter of the Association for Computational Linguistics: Human Language
  Technologies, Volume 1 (Long and Short Papers)}, pages 4171--4186, 2019.

\bibitem{antoun2020arabert}
Wissam Antoun, Fady Baly, and Hazem Hajj.
\newblock Arabert: Transformer-based model for arabic language understanding.
\newblock In {\em LREC 2020 Workshop Language Resources and Evaluation
  Conference 11--16 May 2020}, page~9, 2019.

\bibitem{sanh2019distilbert}
Victor Sanh, Lysandre Debut, Julien Chaumond, and Thomas Wolf.
\newblock Distilbert, a distilled version of bert: smaller, faster, cheaper and
  lighter.
\newblock {\em arXiv preprint arXiv:1910.01108}, 2019.

\bibitem{sun2019fine}
Chi Sun, Xipeng Qiu, Yige Xu, and Xuanjing Huang.
\newblock How to fine-tune bert for text classification?
\newblock In {\em China National Conference on Chinese Computational
  Linguistics}, pages 194--206. Springer, 2019.

\bibitem{scikit-learn}
F.~Pedregosa, G.~Varoquaux, A.~Gramfort, V.~Michel, B.~Thirion, O.~Grisel,
  M.~Blondel, P.~Prettenhofer, R.~Weiss, V.~Dubourg, J.~Vanderplas, A.~Passos,
  D.~Cournapeau, M.~Brucher, M.~Perrot, and E.~Duchesnay.
\newblock Scikit-learn: Machine learning in {P}ython.
\newblock {\em Journal of Machine Learning Research}, 12:2825--2830, 2011.

\bibitem{Flair_akbik2018coling}
Alan Akbik, Duncan Blythe, and Roland Vollgraf.
\newblock Contextual string embeddings for sequence labeling.
\newblock In {\em {COLING} 2018, 27th International Conference on Computational
  Linguistics}, pages 1638--1649, 2018.

\bibitem{wolf-etal-2020-transformers}
Thomas Wolf, Lysandre Debut, Victor Sanh, Julien Chaumond, Clement Delangue,
  Anthony Moi, Pierric Cistac, Tim Rault, Rémi Louf, Morgan Funtowicz, Joe
  Davison, Sam Shleifer, Patrick von Platen, Clara Ma, Yacine Jernite, Julien
  Plu, Canwen Xu, Teven~Le Scao, Sylvain Gugger, Mariama Drame, Quentin Lhoest,
  and Alexander~M. Rush.
\newblock Transformers: State-of-the-art natural language processing.
\newblock In {\em Proceedings of the 2020 Conference on Empirical Methods in
  Natural Language Processing: System Demonstrations}, pages 38--45, Online,
  October 2020. Association for Computational Linguistics.

\bibitem{PyTorch_NEURIPS2019_9015}
Adam Paszke, Sam Gross, Francisco Massa, Adam Lerer, James Bradbury, Gregory
  Chanan, Trevor Killeen, Zeming Lin, Natalia Gimelshein, Luca Antiga, Alban
  Desmaison, Andreas Kopf, Edward Yang, Zachary DeVito, Martin Raison, Alykhan
  Tejani, Sasank Chilamkurthy, Benoit Steiner, Lu~Fang, Junjie Bai, and Soumith
  Chintala.
\newblock Pytorch: An imperative style, high-performance deep learning library.
\newblock In H.~Wallach, H.~Larochelle, A.~Beygelzimer, F.~d\textquotesingle
  Alch\'{e}-Buc, E.~Fox, and R.~Garnett, editors, {\em Advances in Neural
  Information Processing Systems 32}, pages 8024--8035. Curran Associates,
  Inc., 2019.

\end{thebibliography}

\end{document}